\documentclass{article}

\usepackage{amsmath, amssymb}           % maths formatting and symbols
\usepackage{graphicx}                   % insert graphics
\usepackage{pdfpages}
\usepackage{xcolor}
\usepackage{color}
\usepackage[sc]{titlesec}
\usepackage{iclr2017_workshop,times}
\usepackage{float}
\restylefloat{table}
\usepackage{tabularx}
\newcolumntype{Y}{>{\centering\arraybackslash}X}
\usepackage{wrapfig}
\usepackage{natbib}
\usepackage{hyperref}

\title{Early Methods for Detecting\\ Adversarial Images}

\author{Dan Hendrycks\thanks{Work done while the author was at TTIC. Code available at \href{https://github.com/hendrycks/fooling
}{github.com/hendrycks/fooling
} }\\University of Chicago\\\texttt{dan@ttic.edu} \And Kevin Gimpel\\ Toyota Technological Institute at Chicago\\\texttt{kgimpel@ttic.edu}}

\begin{document}

\date{}
\maketitle
\begin{abstract}
Many machine learning classifiers are vulnerable to adversarial perturbations. An adversarial perturbation modifies an input to change a classifier's prediction without causing the input to seem substantially different to human perception. We deploy three methods to detect adversarial images. Adversaries trying to bypass our detectors must make the adversarial image less pathological or they will fail trying. Our best detection method reveals that adversarial images place abnormal emphasis on the lower-ranked principal components from PCA. Other detectors and a colorful saliency map are in an appendix.
\end{abstract}

\section{Introduction}
Images can undergo slight yet pathological modifications causing machine learning systems to misclassify, all while humans barely can notice these perturbations. These types of manipulated images are adversarial images \citep{goodfellow}, and their existence demonstrates frailties in machine learning classifiers and a disconnect between human and computer vision.

This unexpected divide can allow attackers complete leverage over some deep learning systems. For example, adversarial images could cause a deep learning classifier to mistake handwritten digits, thereby fooling the classifier to misread the amount on a check \citep{papernot, physical}. Other fooling data could evade malware detectors or spam filters that use a deep learning backend \citep{malware}. Worse, generating adversarial images requires no exact knowledge of the deep learning system in use, allowing attackers to achieve consistent control over various classification systems \citep{transfer, transfer2}.

In this paper, we make progress on detecting adversarial images, and we contribute a new saliency map to make network classification decisions more understandable. We start by presenting our first detection method which reveals that adversarial images place abnormally strong emphasis on principal components which account for little variance in the data. Our second methods uses the observation from \cite{errordetection}, which states that correctly classified and many out-of-distribution examples tend to have different softmax distributions, but now we apply this to adversarial images. For our third detection method, we show that adversarial image reconstructions can be worse than clean image reconstructions if the decoder uses classification information. Attempts by an adversary to bypass our detector either fails or compels the adversary to make larger modifications to a clean image.

\section{Detecting Adversarial Images}

\subsection{PCA Whitening Adversarial Images}
We now show the first of three adversarial image detection methods. For the first detector, we need to PCA whiten or ``sphere'' an input. To do this, we must center the training data about zero, compute the covariance matrix $C$ of the centered data, and find the SVD of $C$ which is $C=U\Sigma V^\mathsf{T}$. We can perform PCA whitening by taking an input example $x$ and computing $\Sigma^{-1/2}U^\mathsf{T}x$, giving us the PCA whitened input. This whitened vector has as its first entry a coefficient for an eigenvector/principal component of $C$ with the largest eigenvalue. Later entries are coefficients for eigenvectors with smaller eigenvalues. Adversarial images have different coefficients for low-ranked principal components than do clean images.

\begin{figure}[!ht]
	\centering
	\noindent\makebox[\textwidth]{\includegraphics[width=\textwidth]{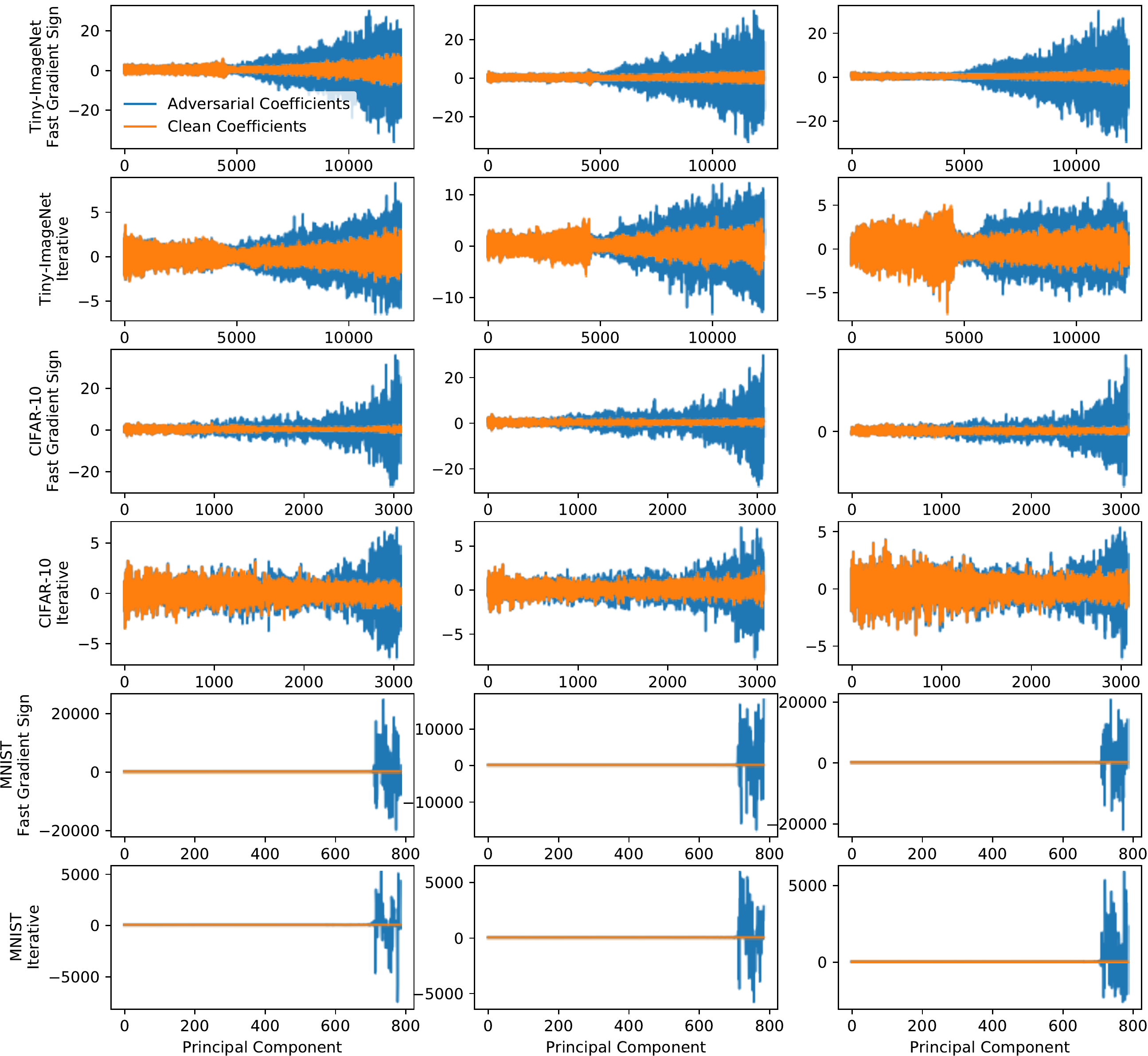}}
	\caption{Adversarial images abnormally emphasize coefficients for low-ranked principal components. Each plot corresponds to coefficients for a randomly chosen clean and adversarial image pair. For concreteness, the 100th ``Principal Component'' Coefficient is the 100th entry of $\Sigma^{-1/2}U^\mathsf{T}x$ for an input image $x$. Examples are randomly chosen.}\label{fig:coefficients}
\end{figure}

Adversarial image coefficients for later principal components have consistently greater variance as shown in Figure \ref{fig:coefficients}. In fact, we use coefficient variance as our sole feature for detecting adversarial images. For this experiment, we try using variance to detect whether Tiny-ImageNet, CIFAR-10, and MNIST images had adversarial perturbations applied. To reduce the chance of bugs in this experiment, we used a pretrained Tiny-ImageNet classifier and a popular, pre-existing adversarial image generator from github \citep{cs231n} in addition to our own implementation of an adversarial image generator for CIFAR-10 and MNIST. In the case of Tiny-ImageNet test set images, each pixel is in the range $[0,255]$ before we subtract the mean. For these images, we create fast gradient sign (step size is 10) and iteratively generated (step size is 1) adversarial images. For CIFAR-10 and MNIST test set images, the fast gradient sign adversarial images have a step size of $10/255$ and the iteratively generated images use a step size of $1/255$, all while having an $\ell_2$ regularization strength of $\lambda = 10^{-3}$. These step sizes are small since the data domain has width 1 instead of 255. All adversarial examples are clipped after each descent step so as to ensure that adversarial perturbations are within normal image bounds. Finally, we keep the adversarial image only if the softmax prediction probability is above 50\% for its randomly chosen target class. Now, we compute detector values by computing the variance of the coefficients for clean and adversarial images. These variance values are from a subsection of each whitened input. Specifically, we select the 10000th to final entry of a PCA whitened Tiny-ImageNet input and compute one variance value from this vector subsection. For each CIFAR-10 whitened image, we compute the variance of entries starting at the 2500th entry. For MNIST, we compute the variance of the entries starting at the 700th position. The variances are frequently larger for adversarial images generated with fast gradient sign and iterative methods, as demonstrated in Table \ref{tab:whiten}.

\begin{figure}
	\centering
    \includegraphics[width=0.48\textwidth]{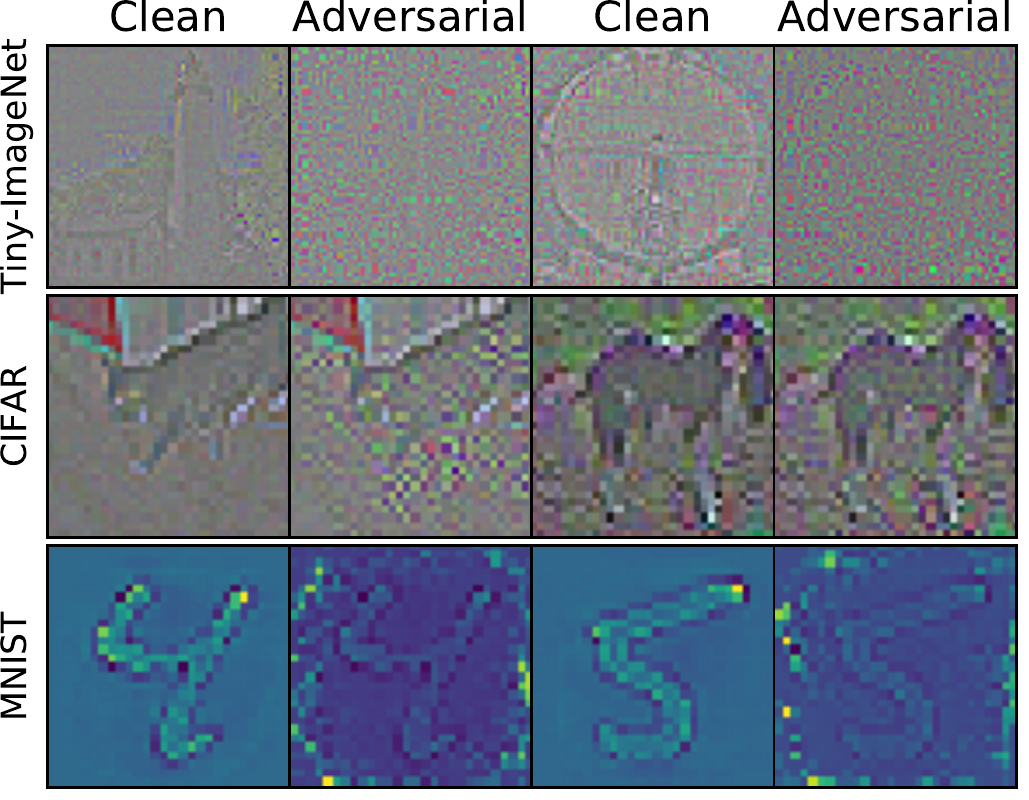}
  	\caption{ZCA whitened adversarial images are often visibly distinct from whitened clean images.}
    \label{fig:whitened}
\end{figure}

\begin{table}
\begin{center}
\begin{tabularx}{\textwidth}{X | *{4}{>{\hsize=.75\hsize}Y}}
\hline 	Dataset & 	Fast Gradient Sign AUROC & Fast Gradient Sign AUPR & Iterative AUROC & Iterative \, AUPR \\ \cline{1-5}
\bf{Tiny-ImageNet} 	& 100.0 & 100.0 & 92.4 & 93.5 \\
\bf{CIFAR-10}		& 100.0 & 100.0 & 92.8  & 91.2 \\
\bf{MNIST}			& 100.0 & 100.0 & 100.0 & 100.0 \\
\hline
\end{tabularx}
\caption{Variance of entries for whitened adversarial images and whitened clean images differ enough to allow reliable detection. Baseline values are 50\%, and ``perfect'' detectors correspond to 100\%. All entries are percentages.}\label{tab:whiten}
\end{center}
\end{table}

We can also try making adversarial images have typical variances, but this seems to fail. In the case of fast gradient sign adversarial images, each image is generated after one large step, the image's coefficient variance explodes with this one large modification. To see this consider the aforementioned coefficient subsections for whitened Tiny-ImageNet, CIFAR-10, and MNIST examples. All fast gradient sign adversarial images have their coefficient variance beyond \emph{ten} standard deviations from the mean coefficient variance for clean images. Remarkably in the case MNIST, 100\% of fast gradient sign images have their coefficient's variance beyond 10 \emph{billion} standard deviations from the mean coefficient variance of clean images. Similarly, iteratively generated adversarial images fail to obtain typical variances. For MNIST, we repeat the iterative generation procedure described earlier but we bound the search procedure. To show how we bound the search, we establish notation. Let $v_\text{example} = \text{Var}((\Sigma^{-1/2}U^\mathsf{T}x_\text{example})_{700:784})$, where $700:784$ is an array subsection from the 700th to final entry. The clean examples $\text{clean1}, \text{clean2}, \ldots$ have each have the variances $v_\text{clean1},v_\text{clean2}, \ldots$. These variances have the group standard deviation $\sigma_\text{clean}=\text{std}([v_\text{clean1},v_\text{clean2},\ldots])$ and mean $\mu_\text{clean}=\text{mean}([v_\text{clean1},\ldots])$. We desire that our adversarial example has its coefficient variance $v_\text{adversarial}$ within one standard deviation $\sigma_\text{clean}$ of the mean coefficient variance for clean examples $\mu_\text{clean}$. This requirement is established by adding the logarithmic barriers $-\log(\mu_\text{clean} + \sigma_\text{clean} - v_\text{adversarial})$ and $-\log(v_\text{adversarial} - \mu_\text{clean} + \sigma_\text{clean})$ to the loss. Under these constraints, the search procedure could not generate any adversarial MNIST digit. We did not try converting these constraints for Tiny-ImageNet because the popular codebase taken from github is substantially different from mainstream libraries. However, if we convert these constraints for CIFAR-10 images, then 92\% of the randomly chosen clean images could not be converted to an adversarial image with satisfactory coefficient variance. In all, encouraging typical coefficient variances appears difficult for fast gradient sign and iterative adversarial images.

% In hindsight, it is reasonable that adversarial perturbations modify the low-ranked principal components. If such perturbations targeted high-ranked principal components, the visible structure of an image would change which we know does not happen. Additionally, since the effect of adversarial perturbations are made manifest by a simple linear whitening transformation, their pathologies do not need to be realized by a specific sequence of nonlinearity and convolutional transformations, and this may help adversarial examples be so transferable \citep{transfer,transfer2}. Moreover, previous works show that adding random Gaussian noise fails to undo an adversarial perturbation, and this is sensible because this noise does not specifically target low-ranked principal components

\section*{Acknowledgments}
We would like to thank Nicolas Papernot and Chris Olah for their helpful feedback on early drafts of this paper. Thank you to Greg Shakhnarovich for numerous insightful suggestions. We would also like to thank the NVIDIA Corporation for donating GPUs used in this research.

\bibliographystyle{iclr2017_conference}
\bibliography{bibliography}

\begin{thebibliography}{20}
\providecommand{\natexlab}[1]{#1}
\providecommand{\url}[1]{\texttt{#1}}
\expandafter\ifx\csname urlstyle\endcsname\relax
  \providecommand{\doi}[1]{doi: #1}\else
  \providecommand{\doi}{doi: \begingroup \urlstyle{rm}\Url}\fi

\bibitem[Davis \& Goadrich(2006)Davis and Goadrich]{auroc}
Jesse Davis and Mark Goadrich.
\newblock The relationship between precision-recall and roc curves.
\newblock In \emph{Proc. of International Conference on Machine Learning
  (ICML)}, 2006.

\bibitem[Dieleman et~al.(2015)Dieleman, SchlÃ¼ter, Raffel, Olson,
  SÃ¸nderby, Nouri, Maturana, Thoma, Battenberg, Kelly, Fauw, Heilman, and
  diogo149]{lasagne}
Sander Dieleman, Jan SchlÃ¼ter, Colin Raffel, Eben Olson, SÃ¸ren~Kaae
  SÃ¸nderby, Daniel Nouri, Daniel Maturana, Martin Thoma, Eric Battenberg,
  Jack Kelly, Jeffrey~De Fauw, Michael Heilman, and Brian~McFee diogo149.
\newblock Lasagne: First release.
\newblock 2015.

\bibitem[Fawcett(2005)]{ROC}
Tom Fawcett.
\newblock \emph{An introduction to ROC analysis}.
\newblock Pattern Recognition Letters, 2005.

\bibitem[Goodfellow et~al.(2015)Goodfellow, Shlens, and Szegedy]{goodfellow}
Ian~J. Goodfellow, Jonathon Shlens, and Christian Szegedy.
\newblock Explaining and harnessing adversarial examples.
\newblock In \emph{International Conference on Learning Representations
  (ICLR)}, 2015.

\bibitem[Grosse et~al.(2016)Grosse, Papernot, Manoharan, Backes, and
  McDaniel]{malware}
Kathrin Grosse, Nicolas Papernot, Praveen Manoharan, Michael Backes, and
  Patrick McDaniel.
\newblock Adversarial perturbations against deep neural networks for malware
  classification.
\newblock \emph{arXiv}, 2016.

\bibitem[Hendrycks \& Gimpel(2016)Hendrycks and Gimpel]{gelu}
Dan Hendrycks and Kevin Gimpel.
\newblock Bridging nonlinearities and stochastic regularizers with {G}aussian
  error linear units.
\newblock 2016.

\bibitem[Hendrycks \& Gimpel(2017)Hendrycks and Gimpel]{errordetection}
Dan Hendrycks and Kevin Gimpel.
\newblock \emph{A Baseline for Detecting Misclassified and Out-of-Distribution
  Examples in Neural Networks}.
\newblock International Conference for Learning Representations (ICLR), 2017.

\bibitem[Kingma \& Ba(2015)Kingma and Ba]{adam}
Diederik Kingma and Jimmy Ba.
\newblock \emph{Adam: A Method for Stochastic Optimization}.
\newblock International Conference for Learning Representations (ICLR), 2015.

\bibitem[Kurakin et~al.(2016)Kurakin, Goodfellow, and Bengio]{physical}
Alexey Kurakin, Ian~J. Goodfellow, and Samy Bengio.
\newblock \emph{Adversarial Examples in the Physical World}.
\newblock arXiv, 2016.

\bibitem[Luo et~al.(2015)Luo, Boix, Roig, Poggio, and Zhao]{foveation}
Yan Luo, Xavier Boix, Gemma Roig, Tomaso~A. Poggio, and Qi~Zhao.
\newblock Foveation-based mechanisms alleviate adversarial examples.
\newblock \emph{arXiv}, 2015.

\bibitem[Manning \& Sch\"{u}tze(1999)Manning and Sch\"{u}tze]{manning}
Chris Manning and Hinrich Sch\"{u}tze.
\newblock \emph{Foundations of Statistical Natural Language Processing}.
\newblock MIT Press, 1999.

\bibitem[Papernot et~al.(2016{\natexlab{a}})Papernot, McDaniel, and
  Goodfellow]{transfer2}
Nicolas Papernot, Patrick McDaniel, and Ian Goodfellow.
\newblock \emph{Transferability in Machine Learning: from Phenomena to
  Black-Box Attacks using Adversarial Samples}.
\newblock arXiv, 2016{\natexlab{a}}.

\bibitem[Papernot et~al.(2016{\natexlab{b}})Papernot, McDaniel, Wu, Jha, and
  Swam]{papernot}
Nicolas Papernot, Patrick McDaniel, Xi~Wu, Somesh Jha, and Ananthram Swam.
\newblock Distillation as a defense to adversarial perturbations against deep
  neural networks.
\newblock In \emph{IEEE Symposium on Security \& Privacy}, 2016{\natexlab{b}}.

\bibitem[Saito \& Rehmsmeier(2015)Saito and Rehmsmeier]{auprbaseline}
Takaya Saito and Marc Rehmsmeier.
\newblock The precision-recall plot is more informative than the roc plot when
  evaluating binary classifiers on imbalanced datasets.
\newblock In \emph{PLoS ONE}. 2015.

\bibitem[Simonyan \& Zisserman(2015)Simonyan and Zisserman]{vgg}
Karen Simonyan and Andrew Zisserman.
\newblock Very deep convolutional networks for large-scale image recognition.
\newblock In \emph{International Conference on Learning Representations
  (ICLR)}, 2015.

\bibitem[Springenberg et~al.(2015)Springenberg, Dosovitskiy, Brox, and
  Riedmiller]{springenberg}
Jost~Tobias Springenberg, Alexey Dosovitskiy, Thomas Brox, and Martin
  Riedmiller.
\newblock Striving for simplicity: The all convolutional net.
\newblock \emph{International Conference on Learning Representations (ICLR)},
  2015.

\bibitem[Szegedy et~al.(2014)Szegedy, Zaremba, Sutskever, Bruna, Erhan,
  Goodfellow, and Fergus]{transfer}
Christian Szegedy, Wojciech Zaremba, Ilya Sutskever, Joan Bruna, Dumitru Erhan,
  Ian Goodfellow, and Rob Fergus.
\newblock Intriguing properties of neural networks.
\newblock \emph{International Conference on Learning Representations (ICLR)},
  2014.

\bibitem[Voss(2015)]{cs231n}
Catalin Voss.
\newblock Visualizing and breaking convnets.
\newblock 2015.
\newblock URL
  \url{https://github.com/MyHumbleSelf/cs231n/blob/master/assignment3/q4.ipynb}.

\bibitem[Zeiler \& Fergus(2014)Zeiler and Fergus]{zeiler}
Matthew~D. Zeiler and Rob Fergus.
\newblock Visualizing and understanding convolutional networks.
\newblock \emph{ECCV}, 2014.

\bibitem[Zhang et~al.(2016)Zhang, Lee, and Lee]{swwae}
Yuting Zhang, Kibok Lee, and Honglak Lee.
\newblock Augmenting supervised neural networks with unsupervised objectives
  for large-scale image classification.
\newblock International Conference on Machine Learning (ICML), 2016.

\end{thebibliography}

\newpage
\appendix

\section{Detector Evaluation}
To justify our three methods for detecting adversarial images, we need to establish our detector evaluation metrics. For detection we have two classes, and the detector outputs a score for both the positive (adversarial) and negative (clean) class. In order to evaluate our detectors, we do not report detection accuracy since accuracy is threshold-dependent, and a chosen threshold should vary depending on a practitioner's trade-off between false negatives (fn) and false positives (fp).

Faced with this issue, we employ the Area Under the Receiver Operating Characteristic curve (AUROC) metric, which is a threshold-independent performance evaluation \citep{auroc}. The ROC curve is a graph showing the true positive rate ($\text{tpr}=\text{tp}/(\text{tp} + \text{fn})$) and the false positive rate ($\text{fpr}=\text{fp}/(\text{fp} + \text{tn})$) against each other. Moreover, the AUROC can be interpreted as the probability that an adversarial example has a greater detector score/value than a clean example \citep{ROC}. Consequently, a random positive example detector corresponds to a 50\% AUROC, and a ``perfect'' classifier corresponds to 100\%.\footnote{A debatable, imprecise interpretation of AUROC values may be as follows: 90\%|100\%: Excellent,\; 80\%|90\%: Good, \; 70\%|80\%: Fair, \; 60\%|70\%: Poor, \; 50\%|60\%: Fail.}

The AUROC sidesteps the issue of selecting a threshold, as does the Area Under the Precision-Recall curve (AUPR) which is sometimes deemed more informative \citep{manning}. One reason is that the AUROC is not ideal when the positive class and negative class have greatly differing base rates, which can happen in practice. The PR curve plots the precision ($\text{tp}/(\text{tp} + \text{fp})$) and recall ($\text{tp}/(\text{tp} + \text{fn})$) against each other. The baseline detector has an AUPR equal to the precision \citep{auprbaseline}, and a ``perfect'' classifier has an AUPR of $100\%$. To show how a classifier behaves with respect to precision and recall, we also show the AUPR.

\section{Additional Experiments}
\subsection{Softmax Distributions of Adversarial Images}
\cite{errordetection} show that many out-of-distribution examples tend to have different softmax output vectors than in-distribution examples. For example, in-distribution examples tend to have a greater maximum probability than softmax vectors for out-of-distribution examples. Similarly, the KL divergence of the softmax distribution $p$ from the uniform distribution $u$ is larger for clean examples than for out-of-distribution examples. Therefore, adversarial examples which stop being modified when the predicted class probability is around 50\% are often easily distinguishable from clean examples since the maximum softmax probability for clean examples tend to be larger than 50\% across many tasks. We can easily detect whether an image is adversarial by considering the resulting softmax distribution for adversarial images.

Softmax distributions metrics can be gamed, but when this is done the adversarial image becomes less pathological. To show this, we create an adversarial image with a softmax distribution similar to those of clean examples by constraining the adversarial image generation procedure. In this experiment, we take correctly classified CIFAR-10 test images and calculate the standard deviation of the KL divergence of the softmax output distribution $p$ from the uniform distribution $u$ over $10$ elements ($\sigma_{\text{KL}[p \| u]}$). We create an adversarial image iteratively, taking a step size of $1/255$ with a regularization strength of $\lambda=10^{-3}$ and clipping the image after each step. We then bound the search procedure with a radius $r$ proportional to $\sigma_{\text{KL}[p \| u]}$ about the mean KL divergence of $p$ from $u$ ($\mu_{\text{KL}[p \| u]}$). This bounding is accomplished by adding logarithmic barriers to the loss. With these bounds, the generator must construct an adversarial image within 1000 steps and with a typical KL divergence. If it succeeds, we have a ``Creation Success.'' In Table \ref{tab:softmax}, we observe that constraining the search procedure with finite radii greatly increases the $\ell_2$ distance. Therefore, to get a typical softmax distribution, the adversarial image quality must decline.

\begin{table}
\begin{center}
\begin{tabularx}{\textwidth}{X | *{4}{>{\hsize=.7\hsize}Y}}
\hline 	Distance from mean KL-Div & $\ell_1$ & $\ell_2$ & $\ell_\infty$ & Creation Success \\ \cline{1-5}
\bf{$r = \infty$} 	 							& 23.4 & 0.43 & 0.081 & 100\% \\
$r = \sigma_{\text{KL}[p \| u]}$				& 32.3 & 1.38 & 0.103 & 83\% \\
$r = \frac{1}{2}\sigma_{\text{KL}[p \| u]}$		& 30.9 & 1.08 & 0.101 & 63\% \\
$r = \frac{1}{4}\sigma_{\text{KL}[p \| u]}$		& 44.2 & 2.82 & 0.153 & 70\% \\
\hline
\end{tabularx}
\caption{Constraints degrade the pathology of fooling images. Value $\sigma$ is a standard deviation.}\label{tab:softmax}
\end{center}
\end{table}

\subsection{Reconstructing Adversarial Images with Logits}
\begin{figure}
	\vspace{-10pt}
	\centering
    	\includegraphics[scale=0.55]{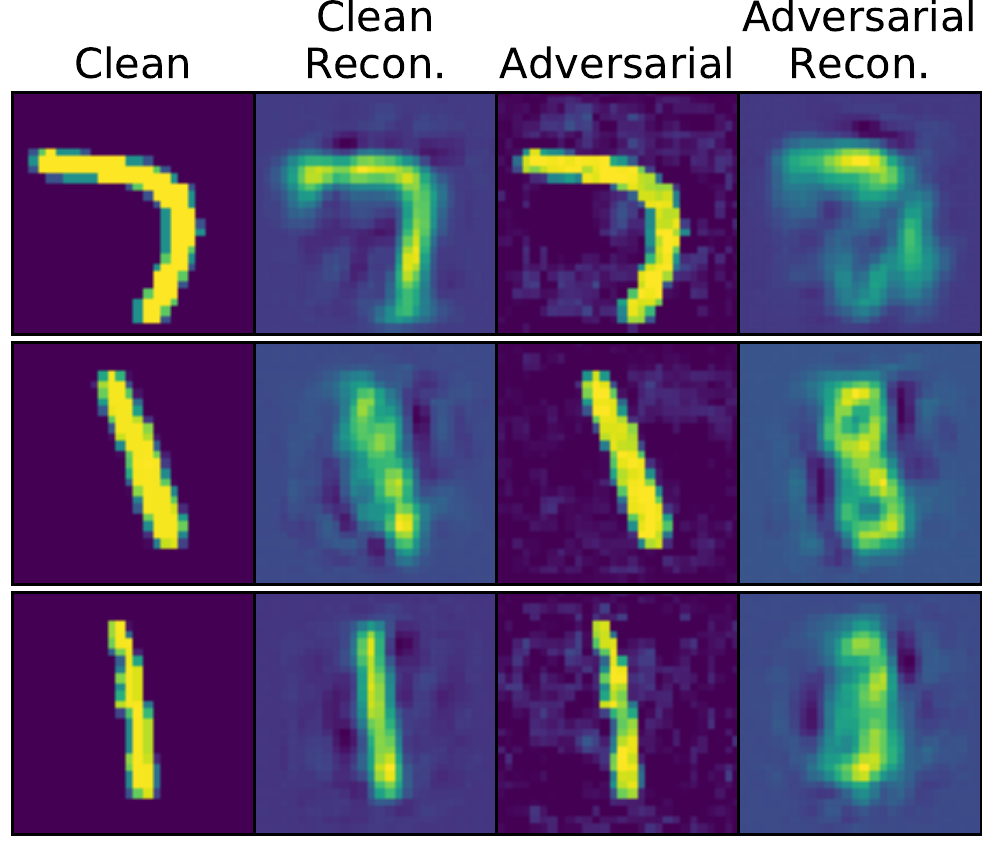}
  	\caption{Adversarial image reconstructions are of lower quality than clean image reconstructions.}\label{fig:recons}
    \vspace{-5pt}
\end{figure}

Our final detection method comes from comparing inputs and their reconstructions. For this experiment, we train an MNIST classifier and attach an auxiliary decoder to reconstruct the input. Auxiliary decoders are sometimes known to increase classification performance \citep{swwae}. We use a hidden layer size of 256, the Adam Optimizer \citep{adam} with suggested parameters, and the GELU nonlinearity \citep{gelu}. The autoencoder bottleneck has only 10 neurons. Crucially, we also feed the logits, viz. the input to a softmax, into the bottleneck layer. After training, we create adversarial images by an iterative procedure like before, except that we increase the $\ell_2$ regularization strength 1000-fold to $\lambda=1$. Now, as evident in Figure \ref{fig:recons}, the reconstructions for adversarial images look atypical. From this observation, we build our final detector. The detector's input score is the mean difference of the input image and its reconstruction. This difference is greater between adversarial images and their reconstruction than for clean examples, so we can detect an adversarial example from a clean example with an AUROC of 96.2\% and an AUPR of 96.6\% where the baseline values are 50\%. Thus the differences between inputs and their reconstructions can allow for adversarial image detection.

\section{A Saliency Map}

Let us now stop considering adversarial images and turn to a different AI Safety goal. This goal is to make neural networks more interpretable. A common way to understand network classifications is through saliency maps. And a simple way to make a saliency map \citep{zeiler} is by computing the gradient of the network's logits with respect to the input and backpropagating the error signal without modifying the weights. Then after the error signal traversed backward through the network, we can display the resulting gradient as an image, and this shows us salient parts of the image. A recently proposed technique to improve saliency maps is guided backpropagation \cite{springenberg}. To understand guided backpropagation, let us establish some notation. Let $f_i^{l+1} = \text{ReLU}(x_i) = \max(0,x_i)$ and $R_i^{l+1} = \partial f^{\text{output}}/\partial f_i^{l+1}$. Now while normally in training we let $R^{l}_i = (f_i^l > 0)R^{l+1}_i$, in guided backpropagation we let $R^{l}_i = (f_i^l > 0)(R^{l+1}_i > 0)R^{l+1}_i$. We can improve the resulting saliency map significantly if we instead let $R^{l}_i = (f_i^l > 0)(R^{l+1}_i > 0)$, leading to our saliency map.\footnote{Note that this technique works for other nonlinearities as well. For example, if we use a Gaussian Error Linear Unit and have $g_i^{l+1} = \text{GELU}(x_i)$, then letting $R^{l}_i = g'(x_i)g'(R^{l+1}_i)$ produces similar saliency maps.}

\begin{figure}
	\centering
	\noindent\makebox[\textwidth]{\includegraphics[scale=0.55]{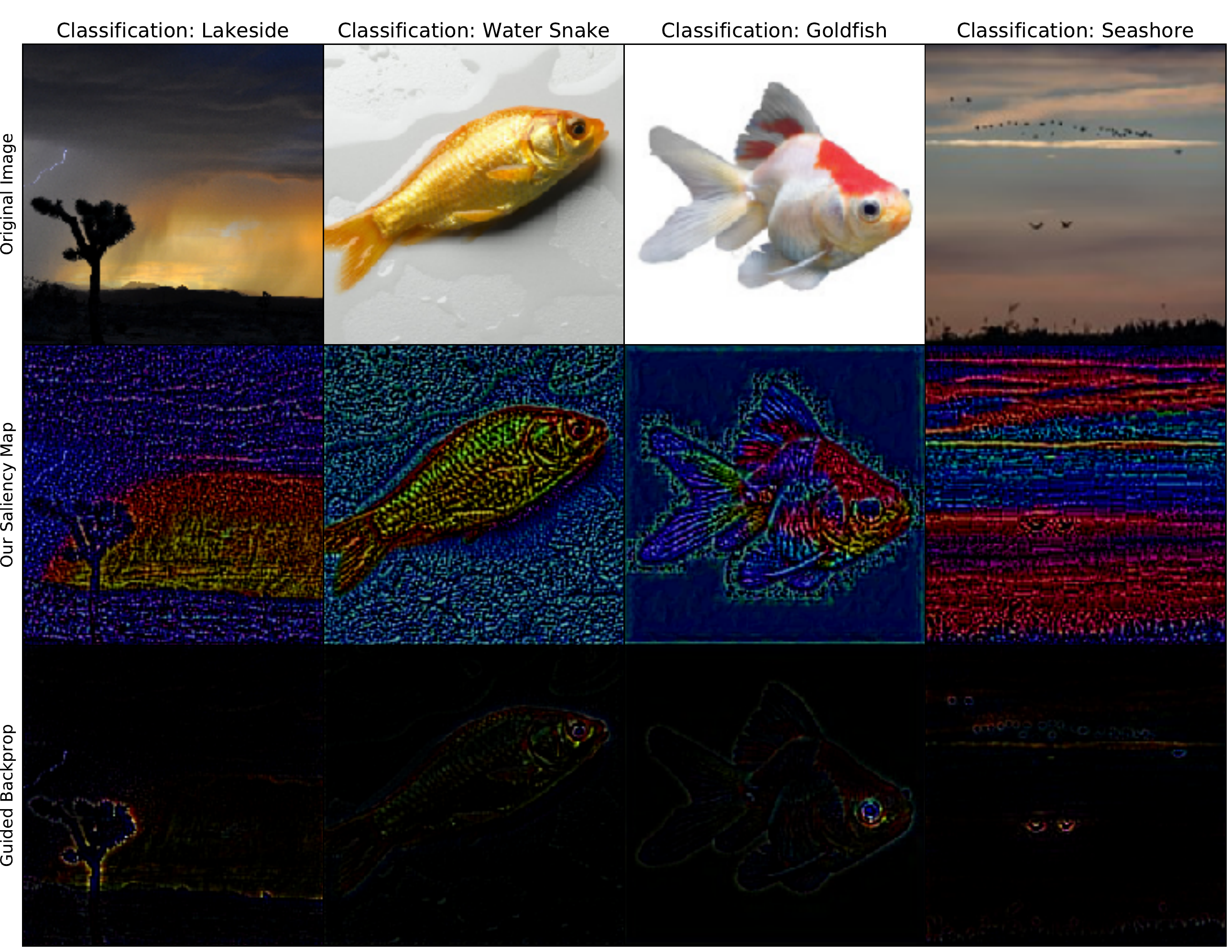}}
	\caption{Our saliency map reveals more saliency details than guided backprogagation.}\label{fig:comparison}
\end{figure}

We demonstrate the results in Figure \ref{fig:comparison} using a pretrained VGG-16 model \citep{lasagne, vgg}. The positive saliency map consists of the positive gradient values. In the first column, the network classifies this desert scene as a lakeside. Our saliency map reveals that coloring the clouds bluer would increase the logits, as would making the sunlit sky more orange. From this we can surmise that the clouds serve as the lakeside water, and the sky the shore. Next, the fish construed as a water snake has a saliency map where the background resembles water, and the fish's scales are greener and more articulated like a water snake. Also with our saliency map, we can see that some white regions of the fish become orange which would increase the logits. Meanwhile, guided backpropagation shows the fish's eye. Last, our saliency map makes it evident that a bluer sky contrasting with a darker lower region would further increase the logits, suggesting the blue sky's role in the image's seashore misclassification. Overall, this new saliency map allows for visualizing clearer sources of saliency.

\section{Discussion and Conclusion}
Although we showed some diverse detectors, future ad-hoc attacks may completely bypass each of these. Indeed, the softmax distribution detector can already be bypassed. But as we saw the adversarial image must contort itself to evade the detector and therefore reduce its level of pathology. We can extend this idea and conclude there may be a better line of attack for stopping adversarial images. Instead of seeking one statistic or one architecture trick to blockade all adversarial attacks, one might use a combination of detectors to more effectively lessen an image's ability to fool. Any narrow metric can be gamed, but this is not necessarily so for an ensemble of several strong but imperfect predictors. \textbf{Therefore, a different avenue to attack adversarial images is by amassing a combination of detectors. These can safeguard against adversarial images by forcing the adversarial perturbation to be more and more conspicuous.}

In this work, we showcased techniques for detecting adversarial images and visualizing salient regions of an image. Future research could add onto this new body of detectors since an ensemble of detectors may be a feasible avenue for defending against adversarial images. A possible future detector could use attention. Say that an adversarial image made a picture of a street classified as a war tank. Since adversarial image attacks seem not to be localized \citep{foveation}, the attention map for the tank's gun turret may be scattered or uniform across the image and thus abnormal and detectable. Other work could be in effective but nearly harmless preprocessing to hinder adversarial perturbations. For example, a preliminary experiment of ours shows that if a CIFAR-10 image is in [0,1], we can square the pixels, apply a slight Gaussian blur, take the square root of the result, and the image often no longer tricks the network allowing us to detect an adversarial attack. If the adversarial generator anticipates this preprocessing through more optimization, the $\ell_2$ norm of the new adversarial image must increase $35\%$ on average. This and other preprocessing techniques could be studied further to add onto an ensemble of defenses. Finally, any general work on out-of-distribution detection or error detection \citep{errordetection} may transfer directly to this adversarial setting. Future research can bear in mind that while adversarial attacks can be invisible to the human eye, we now know that these perturbations can greatly affect the statistics of the underlying image, their reconstruction, or their softmax distribution thereby enabling detection.

\end{document}